\documentclass[letterpaper]{article} 
\usepackage{aaai2026}
\usepackage{times}  
\usepackage{helvet}  
\usepackage{courier}  
\usepackage[hyphens]{url}  
\usepackage{graphicx} 
\usepackage{natbib}  
\usepackage{caption} 
\usepackage{algorithm}
\usepackage{algorithmic}
\usepackage{newfloat}
\usepackage{listings}
\DeclareCaptionStyle{ruled}{labelfont=normalfont,labelsep=colon,strut=off} 
\floatstyle{ruled}
\newfloat{listing}{tb}{lst}{}
\floatname{listing}{Listing}
\usepackage{booktabs}
\usepackage{multirow}
\usepackage{array}

\title{LexChain: Modeling Legal Reasoning Chains for Chinese Tort Case Analysis}

\author {
    Huiyuan Xie\textsuperscript{\rm 1}\equalcontrib,
    Chenyang Li\textsuperscript{\rm 2,3}\equalcontrib,
    Huining Zhu\textsuperscript{\rm 4},
    Chubin Zhang\textsuperscript{\rm 2,3}, \\
    Yuxiao Ye\textsuperscript{\rm 1}\thanks{Corresponding author.}, 
    Zhenghao Liu\textsuperscript{\rm 5}, 
    Zhiyuan Liu\textsuperscript{\rm 1\textdagger}
}
\affiliations {
    \textsuperscript{\rm 1}Tsinghua University\\
    \textsuperscript{\rm 2}Queen Mary University of London\\
    \textsuperscript{\rm 3}Beijing University of Posts and Telecommunications\\
    \textsuperscript{\rm 4}East China University of Political Science and Law\\
    \textsuperscript{\rm 5}Northeastern University\\
    \{xieh,yeyuxiao,liuzy\}@tsinghua.edu.cn~~chenyang.li@se22.qmul.ac.uk
}

\begin{document}

\maketitle

\begin{abstract}
Legal reasoning is a fundamental component of legal analysis and decision-making. Existing computational approaches to legal reasoning predominantly rely on generic reasoning frameworks such as syllogism, which do not comprehensively examine the nuanced process of legal reasoning. Moreover, current research has largely focused on criminal cases, with insufficient modeling for civil cases. In this work, we present a novel framework to explicitly model legal reasoning in the analysis of Chinese tort-related civil cases. We first operationalize the legal reasoning process in tort analysis into the three-module LexChain framework, with each module consisting of multiple finer-grained sub-steps. Informed by the LexChain framework, we introduce the task of tort legal reasoning and construct an evaluation benchmark to systematically assess the critical steps within analytical reasoning chains for tort analysis. Leveraging this benchmark, we evaluate existing large language models for their legal reasoning ability in civil tort contexts. Our results indicate that current models still fall short in accurately handling crucial elements of tort legal reasoning. Furthermore, we introduce several baseline approaches that explicitly incorporate LexChain-style reasoning through prompting or post-training. The proposed baselines achieve significant improvements in tort-related legal reasoning and generalize well to related legal analysis tasks, demonstrating the value of explicitly modeling legal reasoning chains to enhance the reasoning capabilities of language models.
\end{abstract}

\begin{links}
    \link{Data and code}{https://github.com/thunlp/LexChain}
\end{links}

\section{Introduction}

\begin{figure}[t]
    \centering
    \includegraphics[width=0.95\linewidth]{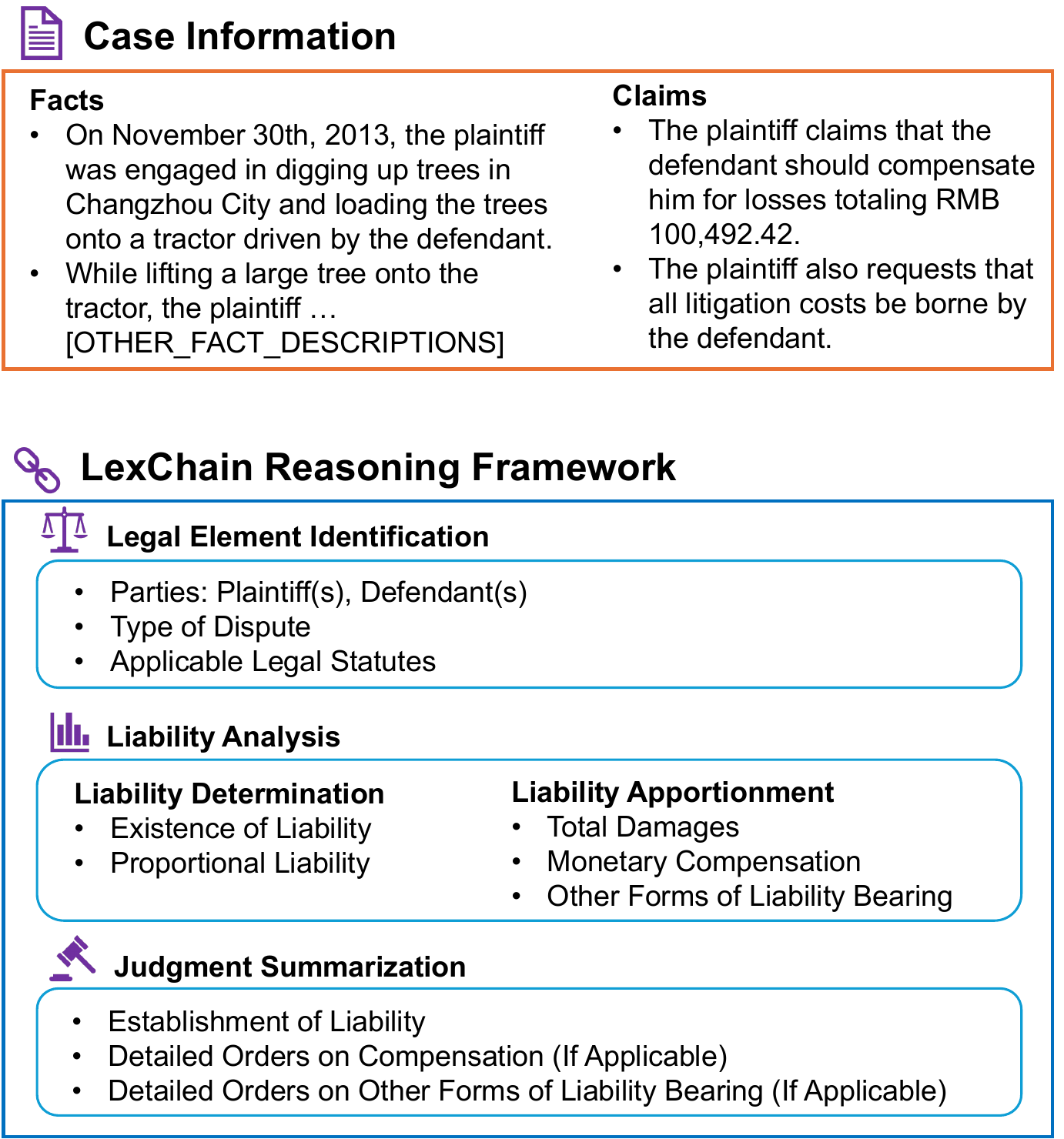}
    \caption{Illustration of the case information (English translation) in a Chinese tort case and the three-module LexChain reasoning framework.}
    \label{fig:lexchain}
\end{figure}

The rapid development of large language models (LLMs) has led to widespread adoption of these tools across diverse domains. Recent models, such as Gemini-2.5-Pro~\cite{comanici2025gemini} and Qwen-3~\cite{yang2025qwen_3}, now position reasoning as a default and essential component of their architecture. While notable progress has been made in advancing the reasoning abilities of LLMs in areas such as mathematics and coding~\cite{wei2022chain,huang2023codecot,el2025competitive}, LLMs' reasoning capabilities in the legal domain, where nuanced outputs are dependent on structured and domain-specific legal reasoning, remain underexplored.

Legal reasoning is a foundational element of legal analysis and decision-making. Legal scholars argue that there is, descriptively, something unique that can be characterized as ``thinking like a lawyer''~\cite{samuelson1997introducing,schauer2009thinking}. This encompasses distinctive forms of reasoning, such as the systematic application of codified legal rules, the decomposition of legal questions into granular elements for structured judgment, and ensuring that similar cases are treated uniformly. Better modeling of legal reasoning not only improves the accuracy and efficiency of legal analysis, but also enhances the interpretability of results by providing explicit reasoning chains.

However, existing work on computational legal reasoning has primarily focused on generic legal reasoning frameworks, such as syllogism~\cite{deng2023syllogistic,jiang2023legal} and IRAC~\cite{yu2023exploring,kuppa2023chain,servantez2024chain}, often simulating reasoning through simple prompting strategies. In addition, research has concentrated predominantly on criminal law, with insufficient attention paid to civil cases, especially tort cases, which are highly relevant to everyday life. Tort law plays a fundamental role in modern legal systems, in both protecting individual rights and compensating plaintiffs for harms suffered.

In this work, we explicitly model legal reasoning chains for tort-related civil cases within the context of Chinese civil law. Drawing on established legal theory~\cite{smith1911legal,moore1999causation}, we operationalize the analytical process of tort adjudication into the LexChain reasoning framework\textemdash a multi-module reasoning chain in which each module comprises a set of key steps or elements essential to tort legal reasoning (see Figure~\ref{fig:lexchain} for an illustration).

We introduce a novel task of tort legal reasoning and construct a benchmark designed to assess whether existing models can correctly identify the key steps and elements required for robust legal reasoning in tort disputes. We evaluate several leading LLMs on this benchmark, observing that accurately modeling the tort reasoning chain remains a challenging task for current models.

Leveraging our LexChain reasoning framework, we further implement reasoning-enhanced baselines using approaches such as legal prompting, supervised fine-tuning (SFT) and direct preference optimization (DPO). Results demonstrate that integrating explicit legal reasoning chains through these techniques consistently improves model performance on identifying key elements for tort case analysis, underscoring the value of explicit legal reasoning modeling in enhancing LLMs' legal reasoning abilities.

Finally, we evaluate our fine-tuned baselines on related legal AI tasks, including Legal Named-Entity Recognition and Criminal Damages Calculation~\cite{fei2024lawbench}. Our reasoning-enhanced models achieve comparable or superior performance to zero-shot counterparts on these tasks, demonstrating the generalizability of reasoning-enhanced approaches to other legal AI tasks.

The contributions of this work are summarized as follows:
\begin{itemize}
    \item We operationalize legal reasoning in the domain of civil tort law by proposing a legally-informed, fine-grained reasoning framework that reflects the structured analytical processes employed in real-world tort litigation.
    \item We introduce the novel task of tort legal reasoning and construct a dedicated benchmark to evaluate models' performance on this task. To the best of our knowledge, the constructed benchmark is the first evaluation dataset that focuses on the assessment of legal reasoning abilities in the context of Chinese civil tort law. In addition, we curate a reasoning-enhanced training dataset to enable effective learning of legal reasoning patterns.
    \item Our empirical results demonstrate that current LLMs exhibit notable limitations in tort-related legal reasoning, particularly in labor dispute cases. Furthermore, we find that generic syllogistic reasoning frameworks do not yield meaningful improvements in this domain, underscoring the need to move beyond generic approaches and adopt domain-specific reasoning paradigms tailored to the complexity of civil tort law. We further show that incorporating LexChain-style reasoning through prompting, SFT and DPO leads to substantial performance gains on the tort reasoning task. These improvements also generalize effectively to other related tasks, underscoring the value of explicitly modeling structured legal reasoning in enhancing LLM performance on legal analysis tasks. 
\end{itemize}

\section{Related Work}

\subsection{Legal Reasoning and Tort Cases}
Legal reasoning refers to the analytical process by which legal practitioners integrate legal norms with case facts to reach a legal conclusion~\cite{samuelson1997introducing,schauer2009thinking}. Legal reasoning holds a central role in judicial practice, serving as the foundation for legal analysis and as a mechanism to ensure the consistency of judicial decisions. Distinct from general forms of logical reasoning, legal reasoning is shaped by the constraints of legal systems, precedential rules and various institutional factors~\cite{scharffs2004character}. The diversity and complexity of real-world cases also make the practical application of legal reasoning highly challenging. 

Tort liability is a form of civil liability that aims to remedy infringed civil rights and safeguard the legitimate interests of civil subjects~\cite{smith1911legal}. In judicial practice, tort disputes are of particular significance, as they encompass a broad spectrum of harms, such as personal injury, property damage and online harassment, permeating many aspects of everyday life~\cite{latin1985problem,wagner2006comparative}. In comparison to other areas of law, tort liability cases exhibit distinctive patterns in their reasoning pathways. A foundational step in tort analysis is the determination of liability, which involves a systematic examination of the constituent elements: conduct, harm, causation and fault~\cite{smith1911legal,catala1965delict,moore1999causation}. Once liability is established, legal practitioners must further consider the nature of the tortious act and the severity of the harm to determine the appropriate forms of liability, with monetary compensation being the most prevalent remedy~\cite{van2001principle}. In certain cases, additional complexities such as compensation for emotional distress and the application of rules for loss offset may arise~\cite{pearson1979apportionment,kontorovich2001mitigation}, demanding complex legal reasoning to successfully adjudicate.

\subsection{Computational Modeling of Legal Reasoning}
Computational modeling of legal reasoning aims to simulate the multi-step inferential processes underpinning legal decision-making~\cite{deng2023syllogistic,fernandes2025llama}. In recent years, structured prompting techniques~\cite{yu2022legal,jiang2023legal} have garnered increasing prominence. A significant line of work leverages syllogism-based prompting~\cite{jiang2023legal}, which encourages models to organize their reasoning in a triadic structure: major premise (legal norm), minor premise (case facts) and conclusion (judgment). Other studies have incorporated schema-based reasoning frameworks such as IRAC~\cite{yu2023exploring,kuppa2023chain}, encouraging generated outputs to resemble authentic legal argumentation. Another logic-based approach~\cite{servantez2024chain} focuses on atomistic parsing of legal rules, decomposing complex provisions into atomic Boolean conditions, allowing for fine-grained alignment with statutory interpretation practices.

In addition to prompting methods, supervised fine-tuning has been used to integrate legal reasoning patterns into models. For instance, syllogism-based fine-tuning~\cite{deng2023syllogistic} encodes syllogistic legal deduction into models, reinforcing deductive reasoning during inference. Similarly, schema-based fine-tuning~\cite{fernandes2025llama} trains models to transform case analyses into structured IRAC representations, effectively internalizing schema-based scaffolds within models' reasoning process.

\begin{figure}[t]
    \centering
    \includegraphics[width=0.99\linewidth]{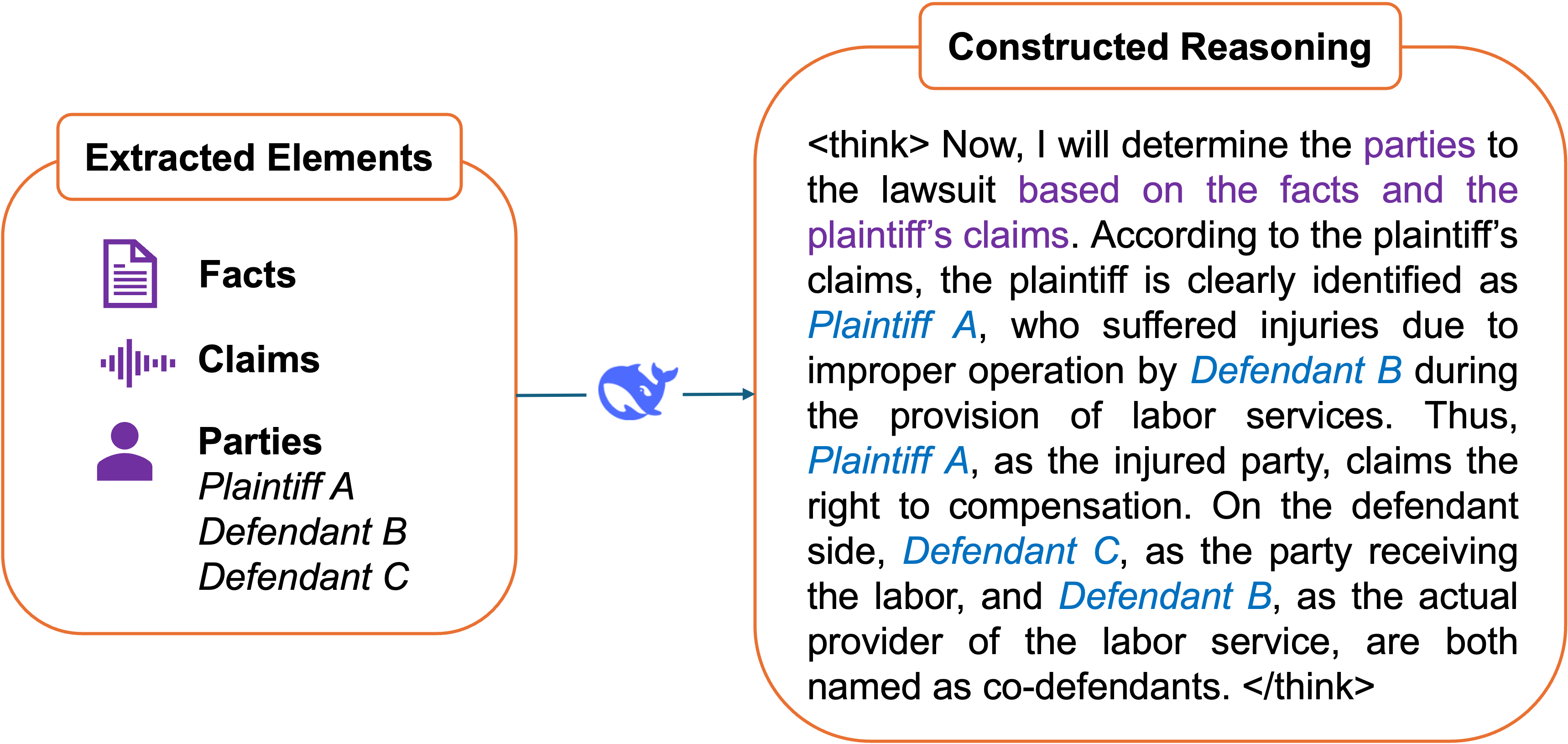}
    \caption{An example of the constructed reasoning text for identifying plaintiff(s) and defendant(s) based on legal elements extracted from the original judicial document.}
    \label{fig:cot_example}
\end{figure}

\section{LexChain: Benchmarking Legal Reasoning}

To model legal reasoning in tort law, we first conceptualize tort-related decision-making as a structured, multi-step reasoning process, which we further operationalize as a legal reasoning chain, referred to here as LexChain. Stemming from this conceptual base, we propose a novel task of tort legal reasoning and construct an evaluation dataset (LexChain$_{eval}$). We further construct a reasoning-enhanced training dataset (LexChain$_{train}$) that incorporates explicit LexChain-style reasoning tracing, which can be used to fine-tune models for improved legal reasoning.

\subsection{Operationalization of Legal Reasoning Chains}
\label{subsec:chain}

Tort analysis exhibits distinctive reasoning structures, particularly centered around the establishment and apportionment of liability. The analytical process typically begins with determining whether liability exists, and proceeds to assess the severity and consequences of the tortious act in order to determine the extent of legal responsibility and appropriate forms of compensation.

Drawing on established legal frameworks~\cite{smith1911legal,catala1965delict,van2001principle} and consultations with practicing judges and lawyers, we formalize tort analysis into a structured, three-module reasoning chain. Each module is further decomposed into a list of specific legal sub-tasks, operationalizing the abstract process of legal reasoning into concrete, verifiable components suitable for computational modeling and evaluation.

The LexChain reasoning framework consists of three modules: (1) a legal element identification module, which identifies relevant foundational information necessary for legal analysis, (2) a liability analysis module, the core module, which centers on determining and apportioning liability, and (3) a judgment summarization module, which synthesizes the outcomes of the preceding modules into a final legal decision. An illustration of the LexChain reasoning framework is presented in Figure~\ref{fig:lexchain}.

\textbf{Legal element identification module}. This preparatory module focuses on identifying essential elements required for subsequent legal reasoning, including:

\begin{itemize}
    \item \textbf{Parties}. Identification of the plaintiff(s) and defendant(s), clarifying the legal actors involved in the dispute.
    \item \textbf{Dispute type}. Specification of the tort category (e.g., traffic accident, defamation), which signals the nature of the civil interest allegedly infringed. This classification is critical because legal provisions in tort law are often organized according to the type of protected civil interest.
    \item \textbf{Applicable legal provisions}. Retrieval of the relevant statutory sources, which serve as the legal basis for the reasoning to follow.
\end{itemize}

\textbf{Liability analysis module}. As the central component of the LexChain, this module determines whether the defendant(s) should bear legal liability, and if so, how such liability should be apportioned. The reasoning process is divided into two sub-tasks for each defendant:

\begin{itemize}
    \item \textbf{Liability determination}. This sub-task assesses whether the defendant has committed a legally recognized tort and thus should be held civilly liable. Following legal doctrine, we structure this analysis around four foundational elements of tort liability: conduct, harm, causation and fault~\cite{smith1911legal}. The presence and interaction of these elements are evaluated to determine both the existence and extent (i.e., proportional liability) of a defendant's responsibility.
    \item \textbf{Liability apportionment}. Once liability is established, this sub-task determines the specific forms of legal responsibility to be imposed, in accordance with civil law. Common forms include monetary compensation, the restitution of damages or cessation of infringement~\cite{wan2022civil}. Given that monetary compensation is the predominant remedy in tort law~\cite{van2001principle}, we explicitly add a step of damages calculation where the total loss suffered by the victim is estimated. The total of estimated damages is multiplied by the proportion of liability attributed to each defendant to obtain a precise amount of compensation. Additional remedies, such as restitution or cessation of infringement, are also considered where legally applicable.
\end{itemize}

\textbf{Judgment summarization module}. This final module summarizes the outputs of the previous modules to generate a complete and coherent legal decision. It synthesizes whether a tort has been legally established, the nature and extent of the liability, the remedies to be ordered, and the exact compensation to be awarded. This module mirrors the structure of real-world judicial opinions and serves to evaluate whether models can deliver legally sound and procedurally complete conclusions.

\subsection{Evaluating Tort Legal Reasoning in LLMs}
\label{subsec:eval_construction}
Facilitated by the modular and stepwise structure formalized in the LexChain reasoning framework, we introduce a novel task of tort legal reasoning (TLR) to systematically evaluate LLMs' capacity to perform structured legal reasoning in the context of tort law. We construct an evaluation benchmark, LexChain$_{eval}$, to assess models' ability to identify and reason through the critical components involved in tort case analysis. In addition, we propose an LLM-assisted scoring framework to quantify model performance at each reasoning step, enabling interpretable and modular evaluation.

\subsubsection{The Task of Tort Legal Reasoning}

Tort legal reasoning (TLR) requires a model to correctly identify or predict a structured sequence of key legal elements that together form a coherent reasoning chain for tort adjudication. The LexChain framework provides a principled decomposition of the reasoning process into discrete modules, each associated with legally meaningful sub-tasks. We construct the TLR task by closely aligning the task with this formalization, allowing models to be evaluated across the entire reasoning trajectory, from case setup to judgment formulation.

More specifically, given the facts and claims of a tort case, the TLR task evaluates whether a model can correctly infer the following categories of legal reasoning elements:

\begin{itemize}
    \item Foundational legal entities and references: including identification of the plaintiff(s), defendant(s), classification of the dispute (i.e., type of tort) and relevant legal statutes applicable to the case.
    \item Liability-related reasoning elements: including the structured reasoning pathway based on the four essential constituent factors of liability (i.e., conduct, harm, causation and fault), the determination of whether liability is established and, if so, the specific proportion of liability attributed to each defendant, the appropriate compensation amount and other forms of legal responsibility.
    \item Judgment summary: synthesizing the conclusions from prior reasoning steps into a structured summary.
\end{itemize}

By decomposing the evaluation of legal reasoning into verifiable sub-components, the TLR task enables a fine-grained evaluation of LLMs' capacity to simulate human-like legal reasoning in tort contexts.

\subsubsection{Constructing Evaluation Dataset}

We construct the evaluation dataset LexChain$_{eval}$ based on real-world judicial documents. Authentic judicial decisions offer a realistic, diverse source of tort-related legal scenarios. In particular, cases that proceed to litigation typically involve greater factual and legal complexity, as they often center on issues of conceptual ambiguity or interpretive dispute~\cite{schauer2009thinking}. These characteristics make litigated cases especially well-suited for modeling and evaluating tort legal reasoning. 

We start by sampling 1,000 real-world tort case judgments from~\citet{cjo2013}. To ensure that the evaluation dataset centers on substantive legal reasoning, we exclude cases involving procedural matters, such as voluntary withdrawals, jurisdictional transfers or dismissals on procedural grounds during the sampling process. The resulting dataset comprehensively covers major types of tort disputes, encompassing 45 distinct tort subcategories.

We design an LLM-assisted extraction pipeline to systematically retrieve key legal information from court decision texts. This pipeline extracts essential elements required for tort reasoning, including: plaintiff(s), defendant(s), type of dispute, applicable legal statutes, claims of the plaintiff(s), case facts, total damages, liability determinations and judgment outcomes. For elements commonly expressed in structured formats in court decisions, such as parties, claims, citations of law and judgments, we employ rule-based extraction techniques, including keyword matching and regular expression-based heuristics. For elements requiring holistic understanding and document-wise summarization, such as fact descriptions, liability reasoning and damages calculation, we utilize DeepSeek-V3~\cite{liu2024deepseek_v3} to synthesize relevant information from the court decision texts.

To ensure the quality of data present in LexChain$_{eval}$, we conduct a rigorous manual validation process on the automatically extracted case information. All extracted data are reviewed and refined by a trained legal annotator, under the supervision of a senior researcher with extensive experience in legal AI and annotation workflows. The extraction and manual validation processes ensure a high-quality benchmark for evaluating tort-related legal reasoning.

\subsubsection{LLM-Based Scoring}

To systematically evaluate model outputs on the tort legal reasoning task, we adopt an ``LLM-as-a-Judge'' evaluation paradigm, widely used in recent literature due to its demonstrated effectiveness and scalability for automated evaluation~\cite{zheng2023judging,gu2024survey}. We design a structured and automated scoring process using GPT-4o~\cite{hurst2024gpt_4o} as the evaluation model. Importantly, instead of relying on subjective, holistic assessments, we decompose the evaluation into a series of discrete, well-defined sub-dimensions and prompt GPT-4o to make rule-based, factual judgments for each sub-dimension. This decomposition mitigates the risk of evaluation bias and enhances reliability of LLM-based evaluation, aligning with existing practices in recent studies~\cite{liu2023geval,wang2024dhp,gu2024survey}. 

The evaluation covers seven distinct scoring dimensions, corresponding directly to the core reasoning elements examined in the LexChain$_{eval}$ benchmark. For each test instance, GPT-4o is instructed to evaluate the model's response against gold-standard answers, i.e., reference case elements provided in the LexChain$_{eval}$ dataset. The scoring schema for each reasoning element is defined as follows:

\begin{enumerate}
    \item Plaintiff identification: Assesses whether the model accurately identifies all plaintiffs specified in the case. A score of 0 denotes incorrect identification; 1 indicates partially correct identification (e.g., only some plaintiffs identified in a multi-party case); 2 denotes a correct identification.
    \item Defendant identification: Follows the same scoring rubric as plaintiff identification, evaluating the model's accuracy and completeness in recognizing all defendants.
    \item Type of dispute: Evaluates whether the model correctly classifies the nature of the tort dispute (e.g., ``motor vehicle accident liability''). Full alignment with the reference answer is required for credit.
    \item Legal statutes: Assesses the accuracy of applicable legal statutes predicted by the model (e.g., specific articles from the Civil Code or judicial interpretations). Full score of 2 is awarded for correctly identifying all relevant provisions; partial score of 1 is given for incomplete citations; and a score of 0 indicates no valid references.
    \item Total damages: Evaluates the correctness of the model's estimate of total damages, as compared to the reference computation, with allowances for reasonable numerical deviations or unit differences.
    \item Liability apportionment: Measures the model's ability to correctly identify the liable parties, the proportion of liability, forms of civil responsibility (e.g., compensation, restitution) and the associated legal justifications.
    \item Judgment summary: Compares the model-generated judgment summary to the reference list of case outcomes, as provided in the official ruling. Evaluation is based on an overlap of factual items (e.g., compensation amount, remedies awarded) and quantified using the F1 score.
\end{enumerate}

To assess the robustness and reliability of the LLM-based evaluation, we conduct quality validation where we randomly sample 300 instances from GPT-4o's ratings and ask two independent annotators to review the ratings. For each of GPT-4o's ratings, annotators independently assign a binary judgment: a score of 1 if they consider GPT-4o's evaluation correct, and 0 otherwise. In cases where the two annotators disagree, a senior researcher with expertise in legal AI practices is consulted, and the annotators engage in discussion to reach a consensus. The final agreed-upon labels (0 or 1) are used as the reference for calculating the accuracy of GPT-4o's ratings. All initial disagreements are also recorded to calculate the inter-annotator agreement score to ensure a fair assessment. Based on this validation criterion, GPT-4o achieves an overall rating accuracy of 0.949 on the sampled set. Inter-annotator agreement, measured using Cohen's Kappa~\cite{cohen1960coefficient}, yields a score of 0.619, indicating relatively strong consistency between annotators\footnote{According to~\citet{landis1977measurement}, a Cohen's Kappa score between 0.61 and 0.80 signifies ``substantial agreement''.}.


\subsection{Constructing LexChain-Style Training Data}

To prepare training data enriched with legal reasoning information, we sample approximately 10,600 court decision documents for tort cases. For each case, we first extract key legal elements following the same information extraction practice described in LexChain$_{eval}$ construction. We then leverage DeepSeek-V3~\cite{liu2024deepseek_v3} to assist in the automatic construction of legal reasoning chains, building upon the extracted legal elements. For each step in the reasoning process, we determine the specific types of information required to infer towards the current reasoning objective and construct corresponding reasoning formulas. For example, the identification of litigation parties (i.e., plaintiff(s) and defendant(s)) requires consideration of both the case facts and the claims of the plaintiff(s), while the calculation of compensation details depends on the previously determined liability ratio and the victim's actual losses. An illustrative example is presented in Figure~\ref{fig:cot_example}. Each constructed reasoning chain is enclosed in a $<$think$>$ and $<$/think$>$ tag pair and prefixed to the final answer consisting of the extracted elements, resulting in the LexChain$_{train}$ training data. 

We conduct human validation to assess the quality of the reasoning chains automatically generated by DeepSeek-V3. Two annotators independently evaluate 300 samples of LLM-generated reasoning chains based on two assessment criteria: consistency and coherence. Consistency measures the alignment of the generated reasoning text with the original legal elements from which it is constructed, while coherence evaluates whether the output forms a logically sound and well-structured reasoning narrative. The training data constructed by DeepSeek-V3 obtains a consistency rate of 0.973 and a coherence rate of 0.993. The inter-annotator agreement between the two annotators, measured as the Cohen's Kappa score~\cite{cohen1960coefficient}, is 0.866, indicating a strong agreement between annotators.

\begin{table*}[t]
    \centering
    \small
    \begin{tabular}{l|c|ccccccc}
    \toprule
    \textbf{\quad\quad Model} & \textbf{Overall} & \textbf{Plaintiff} & \textbf{Defendant} & \textbf{~Dispute~} & \textbf{~Statute~} & \textbf{Liability} & \textbf{~~Damages~~} & \textbf{Judgment} \\
    \midrule
    \textbf{GPT-4o} & 41.17 & 97.20 & 87.05 & 9.90 & 10.35 & 18.10 & 20.05 & 18.67 \\
    \quad w/ Prompt$_{LC}$ & 48.83 & 97.40 & 89.15 & 17.60 & 28.70 & 30.30 & 25.75 & 25.75 \\
    \textbf{o3-mini} & 40.22 & 97.00 & 86.70 & 6.40 & 11.65 & 18.45 & 17.55 & 13.57 \\
    \quad w/ Prompt$_{LC}$ & 50.41 & 97.80 & 89.10 & 18.70 & 33.40 & 32.80 & 26.55 & 26.89 \\
    \textbf{Claude-Sonnet-4} & 44.24 & 97.00 & 89.30 & 31.50 & 11.90 & 20.15 & 19.55 & 23.59 \\
    \quad w/ Prompt$_{LC}$ & 52.90 & \textbf{98.65} & 91.05 & 38.40 & 30.90 & 34.20 & 28.20 & 30.36 \\
    \midrule
    \textbf{DeepSeek-V3} & 46.91 & 97.80 & 87.35 & 35.20 & 20.85 & 22.45 & 22.60 & 25.60 \\
    \quad w/ Prompt$_{LC}$ & 54.71 & 97.70 & 89.40 & 45.50 & 35.45 & 35.15 & 29.35 & 36.86 \\
    \textbf{DeepSeek-R1} & 45.36 & 97.50 & 89.35 & 28.70 & 12.45 & 23.30 & 21.25 & 27.94 \\
    \quad w/ Prompt$_{LC}$ & \textbf{60.84} & 98.10 & \textbf{92.50} & 58.40 & \textbf{44.30} & \textbf{41.70} & \textbf{37.80} & \textbf{42.92} \\
    \midrule
    \textbf{Qwen-3-8B} & 48.79 & 97.50 & 88.60 & 29.50 & 30.70 & 26.35 & 22.35 & 24.92 \\
    \quad w/ Prompt$_{LC}$ & 53.30 & 97.40 & 89.15 & 38.00 & 39.85 & 32.45 & 27.60 & 28.71 \\
    \quad w/ SFT$_{Syll}$ & 39.65 & 96.40 & 86.70 & 13.90 & 13.05 & 16.00 & 12.55 & 12.47 \\
    \quad w/ SFT$_{LC}$ & 55.36 & 96.85 & 87.85 & \textbf{59.00} & 43.70 & 29.05 & 26.80 & 36.88 \\
    \quad w/ DPO$_{LC}$ & 51.17 & 96.20 & 88.85 & 36.00 & 41.95 & 25.55 & 21.00 & 30.95 \\
    \textbf{InternLM-3-8B} & 41.61 & 96.10 & 86.10 & 20.00 & 15.70 & 17.70 & 16.15 & 15.80 \\
    \quad w/ Prompt$_{LC}$ & 48.58 & 96.85 & 86.20 & 26.40 & 32.95 & 29.65 & 19.55 & 26.10 \\
    \quad w/ SFT$_{Syll}$ & 40.32 & 95.60 & 82.35 & 14.60 & 20.05 & 18.75 & 11.80 & 12.13 \\
    \quad w/ SFT$_{LC}$ & 44.89 & 92.85 & 74.30 & 49.90 & 38.95 & 17.55 & 11.80 & 17.91 \\
    \quad w/ DPO$_{LC}$ & 42.67 & 92.80 & 76.95 & 25.10 & 36.05 & 20.05 & 9.40 & 16.43 \\
    \textbf{Llama-3.1-8B} & 37.16 & 97.05 & 85.20 & 2.80 & 6.75 & 13.20 & 13.50 & 11.71 \\
    \quad w/ Prompt$_{LC}$ & 40.35 & 96.75 & 84.05 & 5.90 & 19.10 & 18.40 & 14.35 & 13.04 \\
    \quad w/ SFT$_{Syll}$ & 33.61 & 93.70 & 79.75 & 1.40 & 4.70 & 9.65 & 12.45 & 1.48 \\
    \quad w/ SFT$_{LC}$ & 51.35 & 97.10 & 86.75 & 48.00 & 39.60 & 24.15 & 23.50 & 25.95 \\
    \quad w/ DPO$_{LC}$ & 44.57 & 96.55 & 85.20 & 15.70 & 33.30 & 18.85 & 15.75 & 19.84 \\
    \bottomrule
    \end{tabular}
    \caption{Evaluation results on the LexChain$_{eval}$ benchmark. Entries labeled with the model name alone (e.g., GPT-4o) represent the zero-shot inference setting. The variant ``w/ Prompt$_{LC}$'' denotes models prompted using LexChain-style legal reasoning prompts. Variants ``w/ SFT$_{Syll}$'' and ``w/ SFT$_{LC}$'' denote models fine-tuned on syllogism-style and LexChain-style SFT data, respectively. The variant ``w/ DPO$_{LC}$'' refers to models optimized with LexChain-style DPO data.}
    \label{tab:main}
\end{table*}

\section{Experiments and Results}

\subsection{Baselines}
We evaluate current LLMs on LexChain$_{eval}$, including GPT-4o~\cite{hurst2024gpt_4o}, o3-mini~\cite{openai2025openai_o3mini}, Claude-Sonnet-4~\cite{anthropic2025claude_4}, DeepSeek-V3~\cite{liu2024deepseek_v3}, DeepSeek-R1~\cite{guo2025deepseek_r1}, Qwen-3-8B~\cite{yang2025qwen_3}, InternLM-3-8B~\cite{internlm3}, and Llama-3.1-8B~\cite{grattafiori2024llama_3.1}, ensuring broad coverage of model architecture families. We experiment with two inference-only settings for all models, zero-shot and legal prompting, and three training-based settings for open-source models\footnote{Although DeepSeek-R1 and DeepSeek-V3 are open-source models, we experimented via API calls due to computational resource considerations, and did not perform fine-tuning for them.}: SFT with syllogism data in~\citet{deng2023syllogistic}, SFT with LexChain-style reasoning data and DPO with LexChain-style preference data. 

\textbf{Zero-shot}. In the zero-shot setting, models are prompted using task descriptions only, without any examples or additional instructions.

\textbf{Legal prompting} (denoted as \textbf{Prompt$_{LC}$}). This prompting approach incorporates structured LexChain-style reasoning chains, guiding models to produce responses that reflect multi-step legal reasoning. This approach is applicable across LLMs without the need for model training.

\textbf{SFT$_{Syll}$}. This setting involves supervised fine-tuning using syllogism-style legal reasoning data provided in~\citet{deng2023syllogistic}. In this approach, training instances are structured into the syllogistic form: a major premise representing applicable legal rules, a minor premise representing case facts, and a conclusion stating the legal outcome. 

\textbf{SFT$_{LC}$}. Using the curated LexChain$_{train}$ data, we perform supervised fine-tuning on three open-source models, including Qwen-3-8B, InternLM-3-8B and Llama-3.1-8B. 

\textbf{DPO$_{LC}$}. In the DPO setting, we use the LexChain$_{train}$ data as the \textit{chosen} samples. We further generate corresponding \textit{rejected} samples by collecting answers generated by zero-shot versions of three models (Qwen-3-8B, InternLM-3-8B and Llama-3.1-8B) on LexChain$_{train}$. The resulting (\textit{chosen}, \textit{rejected}) sample pairs are used for DPO training. 

\subsection{Experiment Setup}

The experiment setup for the baseline approaches evaluated in this work is described below.

\textbf{Model Inference Settings}. For all inferences involving closed-source models, the temperature parameter is set to 0.7. For open-source models, inference settings are determined according to either the official suggestions from model developers or the defaults specified in each model's configuration file. For example, for the Qwen-3-8B variants, the following parameters are used: temperature is 0.6, top\_k is set to 20 and top\_p is set to 0.95. 

For non-reasoning models, the maximum token size is set to 4096 tokens. For reasoning-enabled models (such as DeepSeek-R1 and o3-mini), the maximum token size is increased to 8000 tokens to accommodate the extended outputs required for generating reasoning traces.

\textbf{SFT Experiments}. Models are fine-tuned with a batch size of 16 and a learning rate of 1e-4 with AdamW optimizer~\cite{loshchilov2017decoupled_adamw}. Low-Rank Adaptation~\cite[LoRA]{hu2022lora} is used for parameter-efficient adaptation. We use 10\% of the data held out for validation; the optimal checkpoint is selected based on validation set performance. The three models, Qwen-3-8B, InternLM-3-8B and Llama-3.1-8B, are trained for 120, 160 and 80 steps, respectively. All SFT experiments are carried out using LlamaFactory~\cite{zheng2024llamafactory} on 8 NVIDIA A800 GPUs.

\textbf{DPO Experiments}. The setting for the DPO experiments is similar to that of SFT. The effective batch size for DPO training is set to 8. For each model, the final DPO checkpoint is chosen based on the training step comparable to the optimal SFT checkpoint to enable a reasonable comparison.

\subsection{Evaluation Results on LexChain$_{eval}$}

The evaluation results for a broad range of state-of-the-art LLMs on LexChain$_{eval}$ are reported in Table~\ref{tab:main}. In the zero-shot setting, most models exhibit limited capability in reasoning over the multiple dimensions of tort legal analysis. The best-performing model without any task-specific prompt or fine-tuning is Qwen-3-8B, achieving an overall score of 48.79. Notably, as one of the most widely used LLMs, GPT-4o achieves only 41.17 overall, underscoring the challenge posed by the tort legal reasoning task.

We observe that the LexChain-style legal prompting approach (Prompt$_{LC}$) leads to consistent performance improvements across all models. The largest gain is observed for DeepSeek-R1, achieving a 34.1\% increase over its zero-shot baseline and yielding the highest overall performance among all tested approaches. These results highlight the effectiveness of explicit, structured legal prompting in eliciting enhanced legal reasoning abilities from current models.

We further examine the impact of fine-tuning with reasoning-enhanced data, comparing models trained on syllogistic reasoning data (SFT$_{Syll}$), LexChain-style reasoning data (SFT$_{LC}$) and LexChain-style preference data (DPO$_{LC}$). We find that SFT$_{Syll}$ yields reduced performance compared to zero-shot baselines for all models. This suggests that generic syllogistic reasoning is insufficient to capture the domain-specific structure and nuances in tort reasoning. In contrast, both SFT$_{LC}$ and DPO$_{LC}$ exhibit consistent performance improvements across models, with significant gains\footnote{Significance is tested through paired
t-test ($\alpha$=0.01).} observed for Qwen-3-8B and Llama-3.1-8B.

The consistent performance gains achieved by Prompt$_{LC}$, SFT$_{LC}$ and DPO$_{LC}$ collectively demonstrate the effectiveness of incorporating structured legal reasoning signals, whether through legal prompting or model post-training, to improve model abilities on complex legal reasoning tasks.

\begin{table}[t]
    \centering
    \small
    \begin{tabular}{lcc}
    \toprule
    \textbf{\quad Model} & \textbf{Legal NER} & \textbf{Crim Damage} \\\midrule
    \textbf{Qwen-3-8B}         & 7.87 & \textbf{91.40} \\
    \quad w/ SFT$_{LC}$      & 19.36 & 84.40 \\
    \quad w/ DPO$_{LC}$      & 23.38 & 80.40 \\
    \textbf{InternLM-3-8B}     & 13.67 & 64.80 \\
    \quad w/ SFT$_{LC}$      & 45.58 & 69.20 \\
    \quad w/ DPO$_{LC}$      & 14.14 & 68.00 \\
    \textbf{Llama-3.1-8B}      & 61.88 & 38.80 \\
    \quad w/ SFT$_{LC}$      & \textbf{63.90} & 48.60 \\
    \quad w/ DPO$_{LC}$      & 60.20 & 48.60 \\
    \bottomrule
    \end{tabular}
    \caption{Evaluation results of baseline models on the Legal Named-Entity Recognition (Legal NER) and the Criminal Damages Calculation (Crim Damage) tasks.}
    \label{tab:generalization}
\end{table}

\begin{figure}[t]
    \centering
    \includegraphics[width=0.99\linewidth]{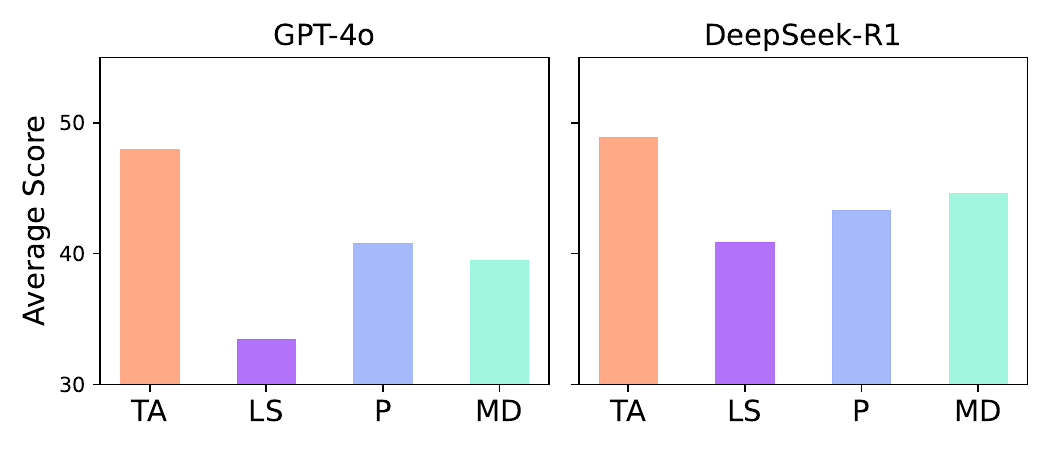}
    \caption{Evaluation results of GPT-4o and DeepSeek-R1 on the four most common types of tort liability, including liability for motor vehicle traffic accident (TA), liability for injury to providers of labor services (LS), liability for product (P) and liability for medical damage (MD).}
    \label{fig:dispute_type}
\end{figure}

\subsection{Generalizability to Other Legal AI Tasks}
To evaluate the generalizability of our proposed reasoning-incorporated approaches to other tasks, we conduct experiments on two related legal AI tasks: Legal Named-Entity Recognition and Criminal Damages Calculation~\cite{fei2024lawbench}. Although both tasks are situated within the context of criminal cases, they share certain similarities with the \textit{legal element identification} and the \textit{compensation calculation} components of our LexChain framework. We compare the performance of reasoning-enhanced models trained using SFT and DPO to their zero-shot counterparts. As shown in Table~\ref{tab:generalization}, both the SFT-based and DPO-based approaches consistently achieve comparable or better performance on these tasks, suggesting that explicitly integrating legal reasoning chains into model training can enhance model performance across a broader spectrum of legal tasks.

\subsection{Analysis of Tort Liability Types}

To analyze model performance across different tort categories, we classify test instances in LexChain$_{eval}$ according to their underlying dispute types. We focus on the four most prevalent types of tort liability in real-world legal practice~\cite{cjo2013}: liability for motor vehicle traffic accident, liability for injury to providers of labor services, liability for product and liability for medical damage, and examine model performance within each category.

Figure~\ref{fig:dispute_type} presents the average overall scores of two representative models, GPT-4o and DeepSeek-R1, on these four categories. As shown, both models achieve their highest scores on traffic accident cases, while their performance drops notably in labor services disputes, the second most common category in real-world tort litigation. In particular, GPT-4o scores below 35 in this category, indicating a significant weakness. This analysis reveals important weaknesses in existing LLMs' legal reasoning capabilities, suggesting a promising direction for future model improvements.

\section{Conclusion}

In this work, we propose LexChain, a structured reasoning framework that explicitly models legal reasoning for Chinese tort-related cases. We introduce a benchmark to evaluate LLMs' ability to identify key reasoning steps in tort case analysis, and find that existing models exhibit notable limitations on this task. By incorporating LexChain-style reasoning through legal prompting, supervised fine-tuning and direct preference optimization, we observe consistent performance improvements on tort reasoning. These reasoning-enhanced models also generalize well to other legal AI tasks. These results highlight the importance of explicitly modeling legal reasoning structures to enhance LLMs' ability for authentic, domain-sensitive legal analysis.

\section*{Acknowledgments}
We thank the anonymous reviewers for their insightful feedback and suggestions. We are deeply grateful to the legal practitioners we consulted in the early stage of this research. Their generous insights and practical perspectives helped shape the legal reasoning framework proposed in this paper. This work is also supported by the AI9Stars community.

\bibliography{aaai2026}

\end{document}